\documentclass[10pt,numafflabel]{elsarticle_LP}
\usepackage{xcolor}
\usepackage{titlesec}
\definecolor{Sepia}{rgb}{.3,.1,.1}
\definecolor{Mahogany}{rgb}{.4,.15,.15}
\definecolor{BrickRed}{rgb}{.6,.2,.2}
\titleformat*{\section}{\LARGE\center\bfseries\itshape\sffamily\color{Sepia}}
\titleformat*{\subsection}{\Large\bfseries\itshape\sffamily\color{Sepia}}
\titleformat*{\subsubsection}{\normalsize\bfseries\sffamily\color{Sepia}} 
\usepackage{graphicx}
\usepackage{lineno}
\modulolinenumbers[5]
\usepackage{setspace}
\usepackage[colorlinks=true]{hyperref}
\usepackage{soul}
\usepackage[labelfont=bf,labelsep=endash,font={sf,small}]{caption}
\usepackage{amssymb}
\usepackage{algorithm}
\usepackage{algpseudocode}
\usepackage{amsmath}
\usepackage{graphics}
\usepackage{epsfig}
\usepackage{bbding}
\usepackage{enumerate}
\usepackage{booktabs}
\usepackage{amsmath}
\usepackage{textcomp}
\usepackage{siunitx}

\usepackage{subfigure}

\usepackage{hyperref}
\hypersetup{
	colorlinks=true,
	linkcolor=blue,
	filecolor=blue,      
	urlcolor=blue,
	citecolor=cyan,
}

\biboptions{authoryear}

\newcommand{\beginsupplement}{%
	\setcounter{table}{0}
	\renewcommand{\thetable}{T\arabic{table}}%
	\setcounter{figure}{0}
	\renewcommand{\thefigure}{S\arabic{figure}}%
}

\begin{document}

\begin{frontmatter}

\title{Quantum Superposition Inspired Spiking Neural Network}


\author[author1,author4,author6]{Yinqian Sun}
\author[author1,author2,author3,author4,author5,author6,author7]{Yi Zeng\corref{mycorrespondingauthor}}\ead{yi.zeng@ia.ac.cn}
\cortext[mycorrespondingauthor]{Yi Zeng}
\author[author1]{Tielin Zhang}
\address[author1]{Research Center for Brain-Inspired Intelligence, Institute of Automation, Chinese Academy of Sciences, Beijing 100190, China}
\address[author2]{Center for Excellence in Brain Science and Intelligence Technology, Chinese Academy of Sciences, Shanghai 200031, China}
\address[author3]{National Laboratory of Pattern Recognition, Institute of Automation, Chinese Academy of Sciences, Beijing 100190, China}
\address[author4]{School of Future Technology, University of Chinese Academy of Sciences, Beijing 100190, China}
\address[author5]{School of Artificial Intelligence, University of Chinese Academy of Sciences, Beijing 100190, China}
\address[author6]{These authors contributed equally}
\address[author7]{Lead contact}



\end{frontmatter}

\singlespacing

\section*{SUMMARY}
Despite advances in artificial intelligence models,  neural networks still cannot achieve human performance, partly due to differences in how information is encoded and processed compared to human brain. Information in an artificial neural network (ANN) is represented using a statistical method and processed as a fitting function, enabling handling of structural patterns in image, text, and speech processing. However, substantial changes to the statistical characteristics of the data, for example, reversing the background of an image, dramatically reduce the performance. Here, we propose a quantum superposition spiking neural network (QS-SNN) inspired by quantum mechanisms and phenomena in the brain, which can handle reversal of image background color. The QS-SNN incorporates quantum theory with brain-inspired spiking neural network models from a computational perspective, resulting in more robust performance compared with traditional ANN models, especially when processing noisy inputs. The results presented here will inform future efforts to develop brain-inspired artificial intelligence.

\section*{INTRODUCTION}
Many machine learning methods using quantum algorithms have been developed to improve parallel computation. Quantum computers have also been shown to be more powerful than classical computers when running certain specialized algorithms, including Shor's quantum factoring algorithm~\citep{shor1999polynomial}, Grover's database search algorithm~\citep{grover1996fast}, and other quantum-inspired computational algorithms~\citep{manju2014applications}.

Quantum computation can also be used to find eigenvalues and eigenvectors of large matrices. For example, the traditional principal components analysis (PCA) algorithm calculates eigenvalues by decomposition of the covariance matrix; however, the computational resource cost increases exponentially with increasing matrix dimensions. For an unknown low-rank density matrix, quantum-enhanced PCA can reveal the quantum eigenvectors associated with the large eigenvalues; this approach is exponentially faster than the traditional method~\citep{lloyd2014quantum}.

K-means is a classic machine learning algorithm that classifies unlabeled datasets into $k$ distinct clusters. A quantum-inspired genetic algorithm for K-means has been proposed, in which a qubit-based representation is employed for exploration and exploitation in discrete ``0'' and ``1'' hyperspace. This algorithm was shown to obtain the optimal number of clusters and the optimal cluster centroids~\citep{xiao2010quantum}. Quantum algorithms have also been used to speed up the solving of subroutine problems and matrix inversion problems~\citep{harrow2009quantum}; for example, Grover's algorithm~\citep{grover1996fast} provides quadratic speedup of a search of unstructured databases. 

The quantum perceptron and quantum neuron computational models combine quantum theory with neural networks~\citep{schuld2014quest}. Compared with the classical perceptron model, the quantum perceptron requires fewer resources and benefits from the advantages of parallel quantum computing~\citep{schuld2015simulating, torrontegui2019unitary}. The quantum neuron model~\citep{cao2017quantum, mangini2020quantum} can also be used to realize classical neurons with sigmoid or step function activation by encoding inputs in quantum superposition, thereby processing the whole dataset at once. Deep quantum neural networks \citep{beer2020training} raise the prospect of deploying deep learning algorithms on quantum computers.   

Spiking neural networks (SNN) represent the third generation of neural network models \citep{maass1997networks} and are biologically plausible from neuron, synapse, network, and learning principles perspectives. Unlike the perceptron model, neurons in an SNN accept signals from pre-synaptic neurons, integrating the post-synaptic potential and firing a spike when the somatic voltage exceeds a threshold. After spiking, the neuron voltage is reset in preparation for the next integrate-and-fire process. SNN are powerful tools for representation and processing of spatial-temporal information. Many types of SNN have been proposed for different purposes. Examples include visual pathway-inspired classification models \citep{zenke2015diverse,zeng2017improving}, basal ganglia-based decision-making models~\citep{herice2016decision,cox2019striatal,zhao2017towards}, and other shallow SNN \citep{khalil2017effects,shrestha2018slayer}. Different SNN may include different types of biologically plausible neurons, e.g., the leaky integrate-and-fire (LIF) model \citep{gerstner2002spiking}, Hodgkin--Huxley model, Izhikevich model \citep{izhikevich2003simple}, and spike response model \citep{gerstner2001framework}. In addition, many different types of synaptic plasticity principles have been used for learning, including spike-timing-dependent plasticity \citep{dan2004spike,fremaux2016neuromodulated}, Hebbian learning \citep{song2000competitive}, and reward-based tuning plasticity \citep{herice2016decision}.

Quantum superposition SNN has theoretic basis in both biology~\citep{Vaziri_2010} and computational models~\citep{kristensen2019artificial}. From one perspective, spiking neuron models, such as the LIF and Izhikevich models, can be reformed by quantum algorithms in order to accelerate their processing using a quantum computer. On the other hand, quantum effects such as entanglement and superposition are regarded as special information-interactive methods
and can be used to modify the classical SNN framework to generate similar behavior to that of particles in the quantum domain. In this work, we follow the latter approach. More specifically, we use a quantum superposition mechanism to encode complementary information simultaneously and further transfer it to spike trains, which are suitable for SNN processing. In our proposed quantum superposition SNN (QS-SNN) model, quantum state representation is integrated with spatio-temporal spike trains in SNN. This characteristic is conducive to good model performance not only on standard image classification tasks but also when handling color-inverted images. QS-SNN encodes the original image and the color-inverted image in the format of quantum superposition; the changing background context demonstrated by the spiking phase and spiking rate contains the image pixels' identity information.

We combine quantum superposition information encoding with SNN for three reasons. First, the possible influence of quantum effects on biological processes and the related quantum brain hypothesis have been theoretically investigated~\citep{Vaziri_2010, FISHER2015593,Weingarten2016New}. Second, quantum superposition states are represented by vectors in complex Hilbert space, in contrast to traditional ANN, which operate in real space only; this is more representative of brain spikes, as the spiking rate and spiking phase spatio-temporal property also have complex number representation. In essence, SNN are more appropriate for quantum-inspired superposition information encoding. Third, current quantum machine learning methods, especially those used for quantum image processing, focus on encoding a classical image in the quantum state, with the image processing methods accelerated by quantum computing~\citep{iyengar2020analysis}. There has been less exploration of the possibility of using a quantum superposition state coding mechanism for different pattern information-processing frameworks. More importantly, owing to the use of statistical methods and fitting functions, current ANN show a huge performance drop when required to recognize a background-inverted image. This inspired us to develop a new information representation method unlike that used in traditional models. The integration of characteristics from SNN and quantum theory is intended to achieve a better representation of multi-states and potentially enable easier solving of tasks that are challenging for traditional ANN and SNN models. 

The subsequent sections describe how complementary superposition information is generated and transferred to spatio-temporal spike trains. A two-compartment SNN is used to process the spikes. The proposed model, combining complementary superposition information encoding with the SNN spatio-temporal property, can successfully recognize a background color-inverted image, which is hard for traditional ANN models.


\subsection*{Complementary superposition information encoding}

\subsubsection*{Quantum image processing}
Quantum image processing combines image processing methods with quantum information theory. There are many approaches to internal representation of an image in a quantum computer, including flexible representation of quantum images (FRQI), NEQR, GQIR, MCQI, and 
QBIP~\citep{iyengar2020analysis, mastriani2020quantum}, which transfer the image to appropriate quantum states for the next step of quantum computing. Our approach is inspired by the FRQI method \citep{le2011FRQI}, as shown in Equations (\ref{FRQI}) and  (\ref{FRQI_limit}):

\begin{equation}
\mathinner{|I(\theta)\rangle}=\frac{1}{2^n}\sum\limits_{i=0}^{2^{2n}-1}(sin(\theta_{i})\mathinner{|0\rangle}+cos(\theta_{i})\mathinner{|1\rangle})\mathinner{|i\rangle},
\label{FRQI}
\end{equation}

\begin{equation}
\theta_{i} \in [0, \frac{\pi}{2}], i= 1, 2, 3, \dots, 2^{2n}-1,
\label{FRQI_limit}
\end{equation}
where $\mathinner{|I(\theta)\rangle}$ is the quantum image, qubit $\mathinner{|i\rangle}$ represents the position of a pixel in the image, and $\theta=(\theta_{0},\theta_{1},\dots,\theta_{2^{2n}-1})$ encodes the color information of the pixels. FRQI satisfies the quantum state constraint in Equation (\ref{FRQI_limit_img}):

\begin{equation}
\parallel\mathinner{|I(\theta)\rangle}\parallel=\frac{1}{2^n}\sqrt{\sum\limits_{i=0}^{2^{2n}-1}(cos^2\theta_{i}+sin^2\theta_{i})}=1.
\label{FRQI_limit_img}
\end{equation}

\subsubsection*{Complementary superposition spikes}
We propose a complementary superposition information encoding method and establish a linkage between quantum image formation and spatio-temporal spike trains. The complement code is widely used in computer science to turn subtraction into addition. We encode the original information and complementary information into a superposition state; one example is shown in Equation (\ref{com_super}), with the rightmost sign bit removed and taking the complement:

\begin{equation}
	\mathinner{|I(\theta_{i})\rangle}=cos(\theta_{i})\mathinner{|[0000001]_b\rangle}+sin(\theta_{i})\mathinner{|[1111110]_b\rangle}.
	\label{com_super}
\end{equation}

\begin{figure}[!htb]
	\centering 
		\includegraphics[width=1.0\linewidth]{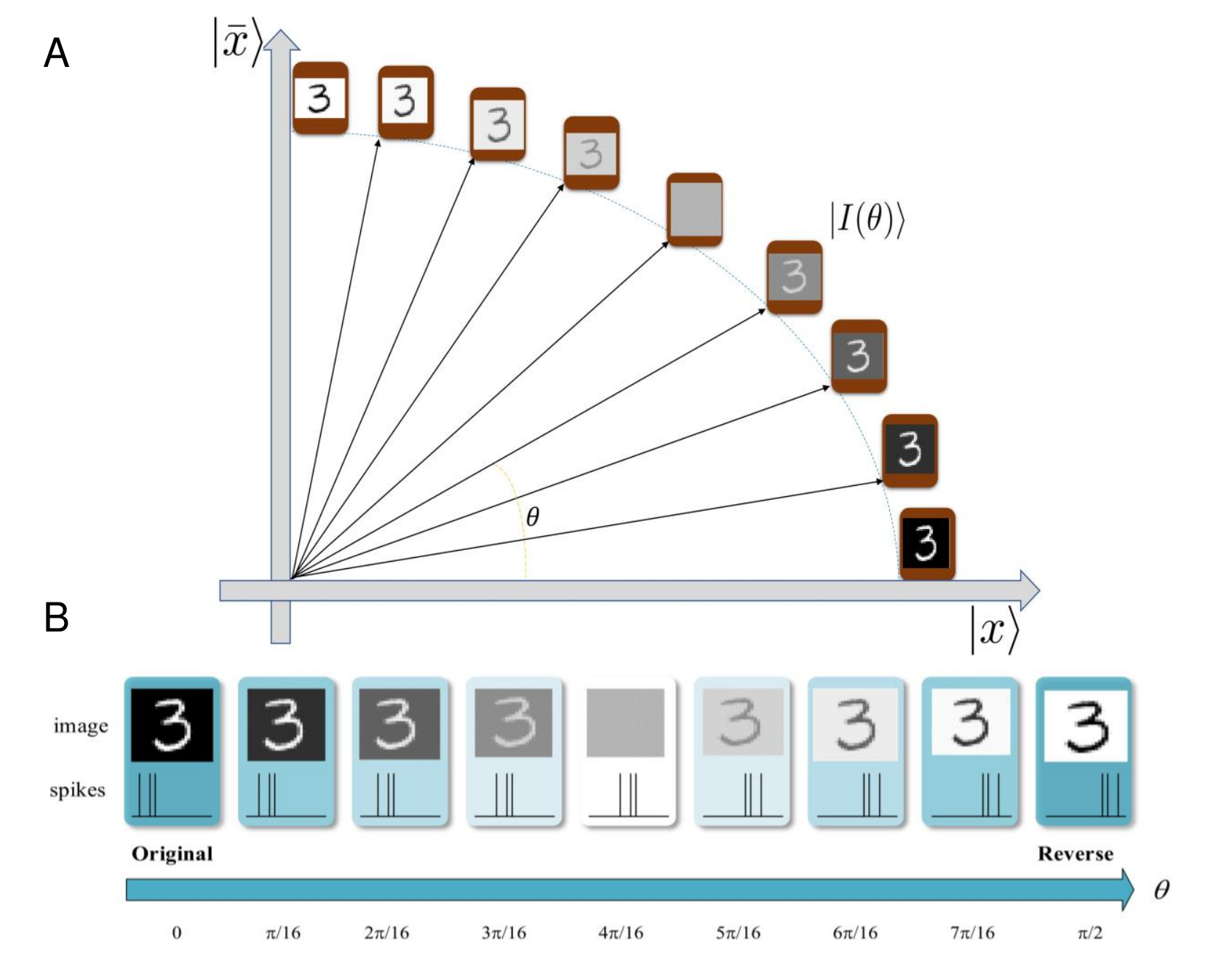}
	\caption{Quantum complementary superposition information encoding. \textbf{(A)} The horizontal axis and longitudinal axis represent $\mathinner{|x\rangle}$ and $\mathinner{|\bar{x}\rangle}$, respectively. The parameter $\theta$ in Equation (\ref{QS-SNN}) measures the degree to which the image background is inverted, from $\theta=0$ (the original image) to the complementary state $\theta=\frac{\pi}{2}$ (totally inverted background).  \textbf{(B)} The top pictures show images inverted to different degrees, and the spikes to which they are encoded. The bottom axis corresponds to the value of $\theta$. It should be noted that the pictures are intuitive demonstration instead of exact display.}
	\label{image_theta_spike}
\end{figure}

Equation (\ref{com_super}) is an illustration of how complementary superposition information encoding works, with no factual significance. In this work, we focus on quantum image superposition encoding, as in Equation (\ref{QS-SNN}). However, it should be noted that any form of information that has a complement format, not just an image, can be encoded as a superposition state. Images in complementary quantum superposition states are further transferred to spike trains, as depicted in Figure \ref{image_theta_spike}. An image in its complementary state has an inverted background. 

\begin{equation}
	\mathinner{|I(\theta)\rangle}=\frac{1}{2^n}\sum\limits_{i=0}^{2^{2n}-1}(cos(\theta_{i})\mathinner{|x_{i}\rangle}+sin(\theta_{i})\mathinner{|\bar{x}_{i}\rangle})\otimes\mathinner{|i\rangle}, \\
	\label{QS-SNN}
\end{equation}

\begin{equation}
	\theta_{i} \in [0, \frac{\pi}{2}], i= 1, 2, 3, \dots, 2^{2n}-1.
	\label{QS-SNN_limitation}
\end{equation}

The complementary quantum superposition encoding is shown in Equations (\ref{QS-SNN}) and (\ref{QS-SNN_limitation}), where the $\mathinner{|i\rangle}$ represent pixel positions. Unlike FRQI, which uses qubits only for color encoding, here we use complementary qubits for encoding both original image pixels $x_{i}$ and the color-inverted image $\bar{x}_{i}$ with $\bar{x}_{i} = 1-x_{i}$, supposing the pixel $x_{i}$ domain ranges from 0 to 1.0. The parameter $\theta_i$ represents the degree of quantum image $\mathinner{|I\rangle}$, mixing the original state $\mathinner{|x\rangle}$ and reverse state $\mathinner{|\bar{x}\rangle}$.

\begin{figure}[!htb]
	\centering		 
		\includegraphics[width=1.0\linewidth]{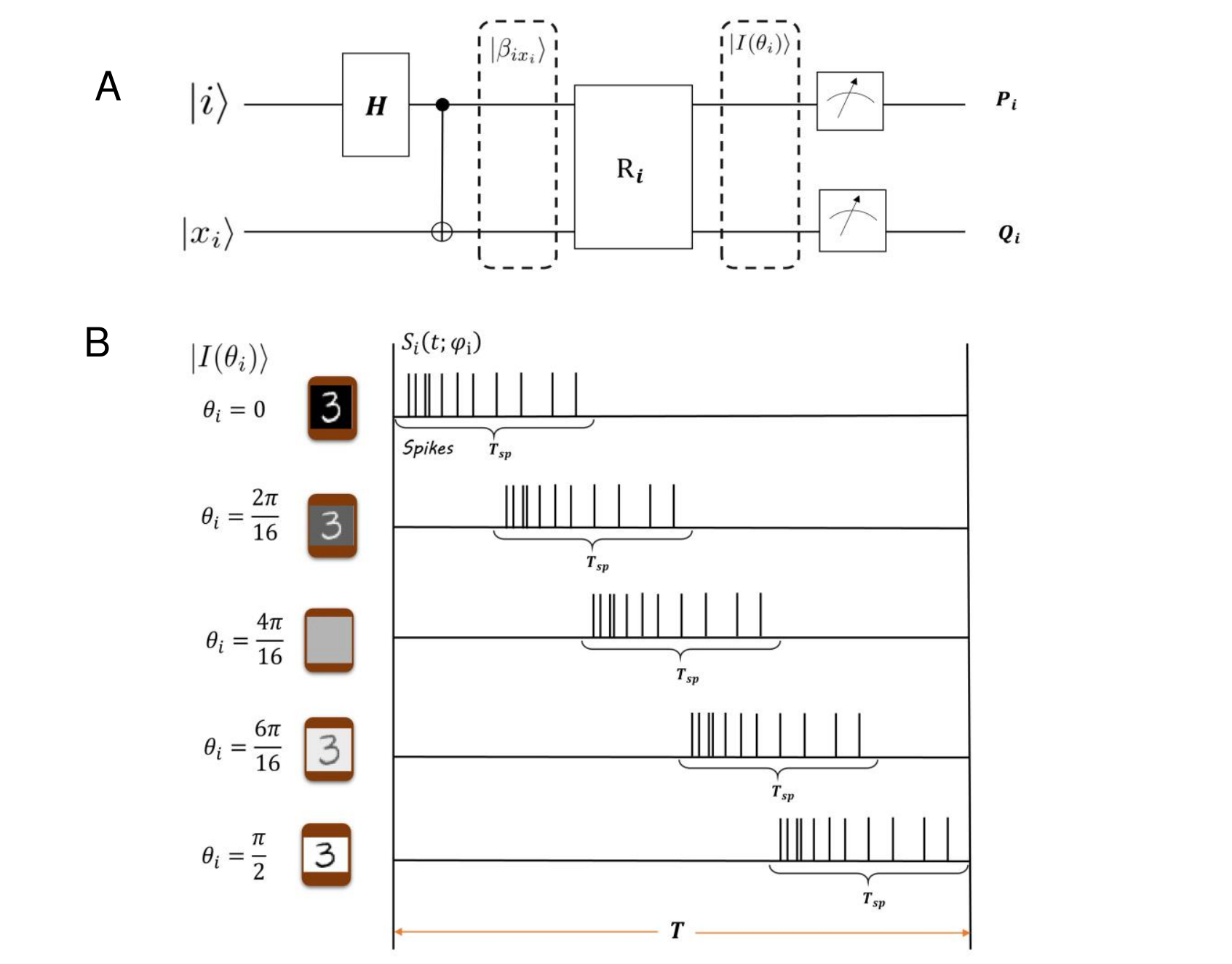}
	\caption{Quantum superposition spike trains. \textbf{(A)} The circuit to generate quantum image. Only one image pixel state is depicted for perspicuity. \textbf{(B)} A schematic diagram shows the transformation of quantum superposition states to spike trains $S_{i}(t;\varphi_i)$. With parameter $\theta_i$ increasing, spike trains are shifted in time dimension.  $T$ is a simulation period in which spikes emerge within the $T_{sp}$ time window. Also, the relation of parameter $\theta_i$ and spiking phase $\varphi_i$ is intuitive example and not exact correspondence.}
	\label{fig_Si}
\end{figure}

We designed quantum circuit for the generation of quantum superposition image $\mathinner{|I(\theta_i)\rangle}$ as shown in the figure~\ref{fig_Si}(A), which is also discussed in~\citep{le2011FRQI, dendukuri2018image}. The quantum state $\mathinner{|x_i\rangle}$ is processed by Hadamard transform $H$ and controlled $NOT$ gate to form the complementary state $\mathinner{|\beta_{ix_i}\rangle}$ with:
	
\begin{equation}
	\mathinner{|\beta_{ix_i}\rangle}=\frac{\mathinner{|0,x_i\rangle}+(-1)^i\mathinner{|1,\bar{x}_i\rangle}}{\sqrt{2}},
	\label{breta_ixi}
\end{equation}

Then rotation matrices $R_i$ is used to encode phase information as

\begin{equation}
	R_i = \left[
	\begin{matrix}
		cos\frac{\theta_i}{2} & -sin\frac{\theta_i}{2} \\ \\
		sin\frac{\theta_i}{2} & cos\frac{\theta_i}{2}
 	\end{matrix}
	\right]
	\label{R_i}
\end{equation}

Finally, the superposition state $\mathinner{|I(\theta_{i})\rangle}$ is measured and two states are retrieved with probability $P_i$ and $Q_i$.

The complex information in quantum encoding is similar to signal processing in SNN. Neuron spikes can encode spatio-temporal information with specific spiking rates and spiking times, which can be used to represent quantum information.

Neuron spikes have the attribute of spatiotemporal dimension, which are identical in shape but differ significantly in frequency and phase, seeing Izhikevich neuron model~\citep{izhikevich2003simple} in Figure S1, and are well-suited to the implementation of the vector form quantum image in Equation (\ref{QS-SNN}). We use spike trains with firing rate $r_{i}$ and firing phase $\varphi_i$ to represent quantum image state $\mathinner{|I(\theta_i)\rangle}$. As shown in figure~\ref{fig_Si}(A), the spike trains containing information of $\mathinner{|I(\theta_i)\rangle}$ can be generated using Equation(\ref{spike_rate}) and Equation (\ref{theta_cal}).



\begin{equation}
	r_{i}=\frac{\parallel\mathinner{|I(\theta)\rangle}\parallel - sin(\varphi_i)}{cos(\varphi_i)-sin(\varphi_i)},
	\label{spike_rate}
\end{equation}

\begin{equation}
	\varphi_i=\mathcal{F}\{arctan(\frac{P_j}{Q_j}) | j =1, 2,...,N.\},
    \label{theta_cal}
\end{equation}


Notation $\mathcal{F}\{X_i\}$ is set operation, and is specific in different tasks. The superposition state encoding $\mathinner{|I(\theta_{i})\rangle}$ is transferred to spike trains $S_i(t;\varphi_i)$, which is generated from a Poisson spikes $S_i(t)$ with spike rate $r_i$ and extended phase $\varphi_i$ shown as Equation~(\ref{spiketrains}). Here, $T$ is the time interval of neuron processing spikes received from pre-synaptic neurons, and $T_{sp}$ is the spiking time window in this period, as shown in Figure~\ref{fig_Si}(B):

\begin{equation}
	S_{i}(t;\varphi_i)=S_i(t-t_0),
	\label{spiketrains}
\end{equation}

\begin{equation}
	t_0 = \frac{\varphi_i}{\pi/2}*(T-T_{sp}).
	\label{phases}
\end{equation}

\subsection*{Two-compartment SNN}

\subsubsection*{Synapses with time-differential convolution}
Synapses play an important part in the conversion of information from spikes in pre-synaptic neurons to membrane potential (or current) in post-synaptic neurons. In this work, the time-differential kernel (TCK) convolution synapse is used, as shown in Equations (\ref{convolution}) and (\ref{postmembrane}) and Figure S2. The spikes, $S_i$, from pre-synaptic neurons are convoluted with a kernel  and then integrated with the dendrite membrane potential $V_{b}(t)$. This process can be considered as a stimulus-response convolution with the form of a Dirac function~\citep{urbanczik2014learning}:

\begin{equation}
	\left\{\begin{array}{l}
		\kappa(t)= \zeta(t) - \zeta(-t) \\
		\zeta(t)=\Theta(t)(e^{-\frac{t}{\tau}})
	\end{array},\right.
	\label{convolution}
\end{equation}

\begin{equation}
	V_{j}^{b}(t) = \sum\limits_{i}w_{i,j}\parallel\int_{-T}^{+T}\kappa(\tau)S_i(\tau)\,d\tau\parallel.
	\label{postmembrane}
\end{equation}

\subsubsection*{Two-compartment neurons}
Both the hidden layer and the output layer contain biologically plausible two-compartment neurons, which dynamically update the somatic membrane potential $V_i(t)$ with the dendrite membrane potential $V^b_{i}(t)$, as shown in  Figure S2.

In the compartment neuron model, $V_i^h(t)$ is the membrane potential of neuron $i$ in the hidden layer, which is updated with Equation (\ref{hidden}); $g_B$, $g_L$, and $\tau_L$ are hyperparameters that represent synapse conductance, leaky conductance, and the integrated time constant, respectively; $V^{PSP}_{j}(t)$ is the synaptic input from neuron $j$; $V_i^{h,b}(t)$ is the dendrite potential with adjustable threshold $b_i^h$ in the hidden layer; and $w_{ij}^h$ is the synaptic weight between the input and hidden layers:

\begin{equation}
	\left\{\begin{array}{l}
		\tau_L\frac{dV_i^{h}(t)}{dt}=-V_i^h(t)+\frac{g_B}{g_L}(V_i^{h,b}(t)- V_i^{h}(t)) \\
		V_i^{h,b}(t) = \sum\limits_{j}w_{ij}^hV^{PSP}_j(t) + b_i^h \\
		V^{PSP}_j(t)=\parallel\int_{-T}^{+T}\kappa(\tau)S_j(\tau)\,d\tau\parallel.
	\end{array}\right.
	\label{hidden}
\end{equation}

The somatic neuron model in the output layer contains 10 neurons corresponding to 10 classes of the MNIST dataset. As shown in Equation (\ref{outputneuron}), the hidden layer neurons deliver signals to the output layer with integrated spike rate $r_i$, which is differentiable; hence, it can be tuned with back-propagation. Here, $V_i^{o}(t)$ is the membrane potential in the output layer, $V_i^{o,b}(t)$ is the dendrite potential, and $r_{max}$ is the hyperparameter for rescaling of fire-rate signals:

\begin{equation}
	\left\{\begin{array}{l}
		\tau_L\frac{dV_i^{o}(t)}{dt}=-V_i^o(t)+\frac{g_B}{g_L}(V_i^{o,b}-V_i^o(t)) \\
		V_i^{o,b}(t) = \sum\limits_{j}w_{ij}^or_{j}(t) + b_i^o \\
		r_j=r_{max}\sigma(V_j^h) \\
		\sigma(x)=1/(1-exp(-x)).
	\end{array}\right.
	\label{outputneuron}
\end{equation}

\begin{figure}[!htb]
	\centering
		\includegraphics[width=1.0\linewidth]{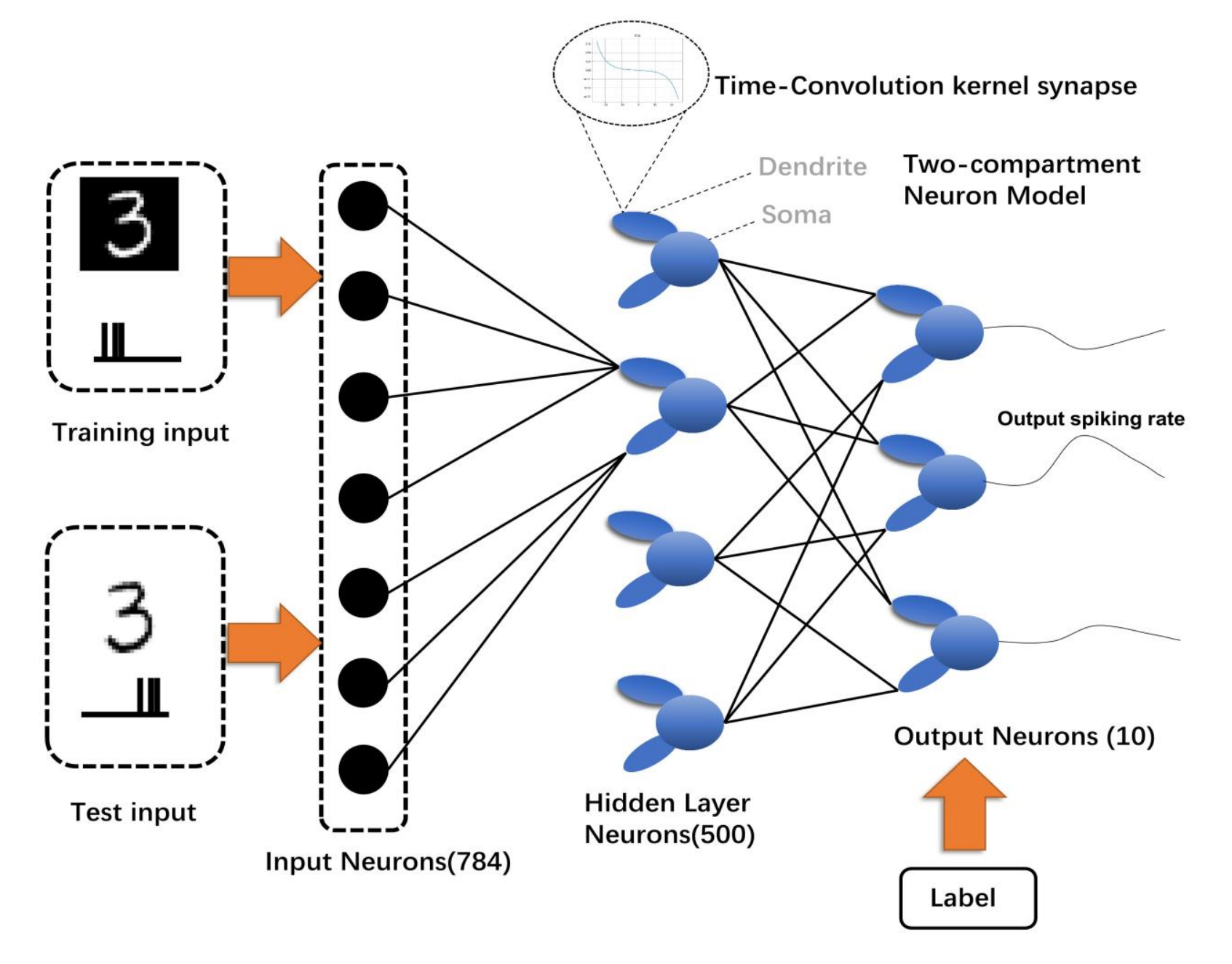}  
	\caption{Quantum superposition SNN, three-layer architecture of QS-SNN with TCK synapses and two-compartment neurons. Images are transferred to spikes as network inputs. The hidden layer is composed of 500 two-compartment neurons with dendrite and soma. The output layer contains 10 two-compartment neurons corresponding to 10 classes. In the training period, only original images are fed to the network, whereas in the test period, the trained network is tested with inverted-background images. Neurons with maximum spiking rates at the output layer are taken as the network prediction and output.}
	\label{QS-SNNarchitecture}
\end{figure}

The shallow three-layered architecture is shown in Figure \ref{QS-SNNarchitecture}. The input layer receives quantum spikes with encoding of complementary qubits. The hidden layer of QS-SNN consists of a two-compartment model with time-differential convolution synapses. Neurons in the output layer are integrated spike-rate neurons, which receive an integrated fire rate from pre-synaptic neurons, as well as the teaching signal  $V_I$ as information about class labels.

\subsection*{Computational experiments}
We examined the performance of the QS-SNN framework on a classification task using background-color-inverted images from the MNIST~\citep{lecun2010mnist} and Fashion-MNIST~\citep{xiao2017Fashion} datasets. QS-SNN encodes the original image and its color-inverted mirror as complementary superposition states and transfers them to spiking trains as an input signal to the two-compartment SNN. The dendrite prediction and proximal gradient methods used to train this model can be found in STAR Methods.

For comparison, we also tested several deep learning models on the color-inverted datasets, including a fully connected ANN, a 10-layers CNN~\citep{LeNet}, VGG~\citep{Simonyan2015VeryDC}, ResNet~\citep{He2016DeepRL} and DeseNet~\citep{huang2017densely}. All models are trained with original image  $x_i$ and then tested on the background reverse image $I(\theta_i)$. The only difference is that, for QS-SNN, quantum superposition state image $\mathinner{|I(\theta_i)\rangle}$ is transferred to spike trains which is compatible with spiking neural networks. In other words, our essential idea is that the spatiotemporal property of neuron spikes enables the brain to transform spatially variant information into time differential information. 

In this work, we formulated the spike trains transformation as the quantum superposition shown in Equation (\ref{QS-SNN}). And we have also demonstrated the numerical calculation of $I(\theta_{i})$ which is used for traditional ANN and CNN model testing in STAR Methods.

In addition, it is worth noting that the superposition state image  $\mathinner{|I(\theta_i)\rangle}$ is constructed from original information $\mathinner{|x_i\rangle}$ and reverse image $\mathinner{|\bar{x}_i\rangle}$. The original image and its complementary reverse information are maintained in the superposition state encoding $\mathinner{|I(\theta_i)\rangle}$ at the same time. Because SNN is not processing pixel value directly, we transformed $\mathinner{|I(\theta_i)\rangle}$ to spike trains, which can be regarded as different expression of superposition state encoding image in spatiotemporal dimension.
\subsubsection*{Standard and color-inverted images}
The standard MNIST dataset contains images of 10 classes of handwritten digits from 0 to 9; images are 28x28 pixels in size, with 60,000 and 10,000 training and test samples, respectively. Fashion-MNIST has the same image size and the same training and testing split as MNIST but contains grayscale images of different types of clothes and shoes.

\begin{figure}[!htb]
	\centering 
		\includegraphics[width=1.0\linewidth]{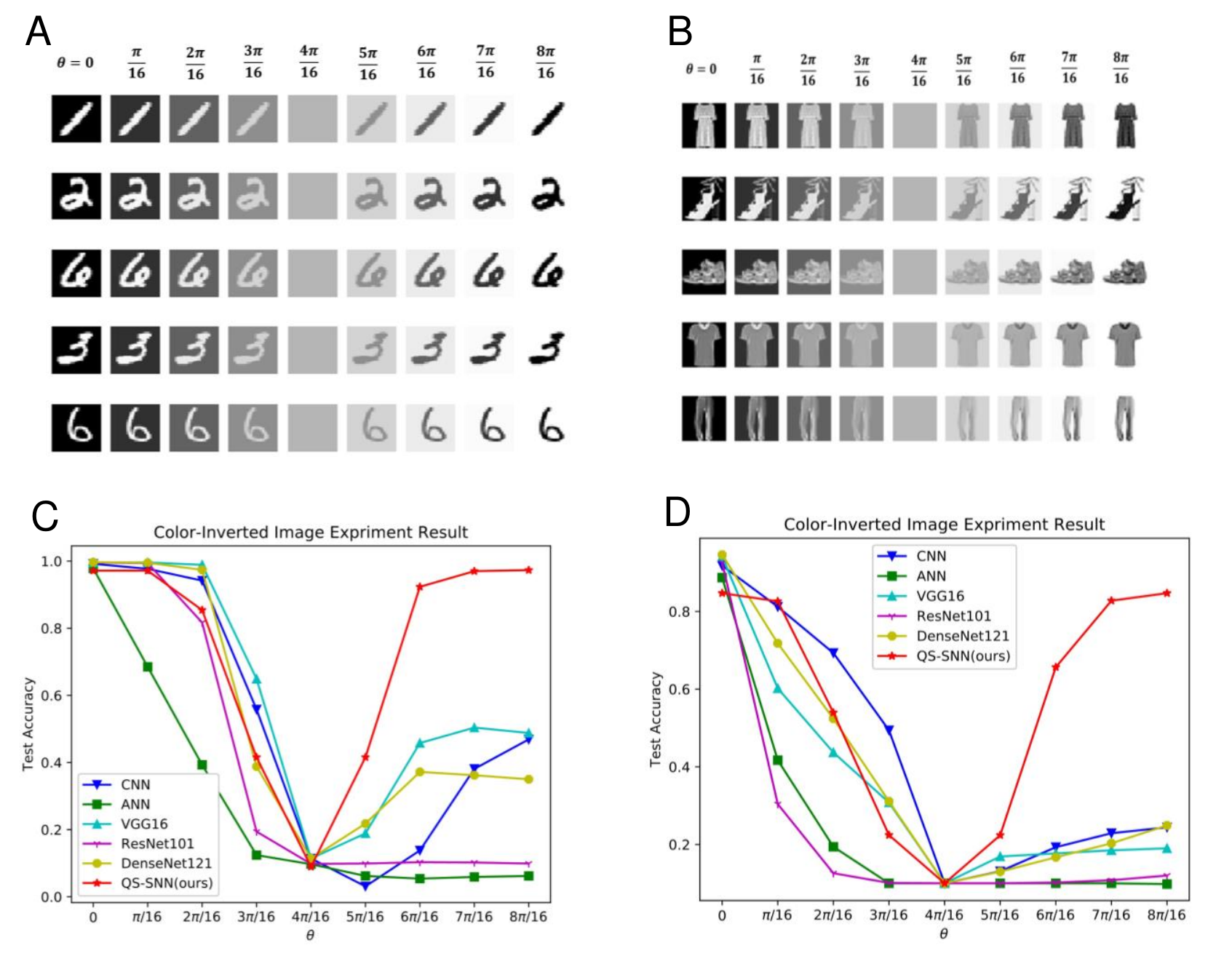}  
	\caption{Classification of color-inverted images. \textbf{(A)} MNIST background color-inverted images. Parameter $\theta$ takes values from 0 to $\frac{\pi}{2}$, denoting the degree of color inversion. \textbf{(B)} Fashion-MNIST background-color-inverted images.  \textbf{(C)} Background color-inverted MNIST classification results. QS-SNN initially showed performance degeneration similar to that of the fully connected ANN and CNN, with $\theta$ values from 0 to $\frac{4\pi}{16}$. However, as the background-inversion degree further increased, QS-SNN gradually recovered its accuracy, whereas the other networks did not. When the background was totally inverted ($\theta=\frac{8\pi}{16}$), QS-SNN showed almost the same performance as when classifying original images, whereas the second-best network (VGG16) retained only half its original accuracy. \textbf{(D)} Background color-inverted Fashion-MNIST results. Similar results as in the MNIST experiment were achieved, with QS-SNN showing an even greater advantage (right-hand side).}
	\label{shift_exp}
\end{figure}

The original MNIST and Fashion-MNIST images and their color-inverted versions, with different degrees of inversion as measured by parameter $\theta$, are depicted in Figure \ref{shift_exp}(A) and (B), respectively.
To be specific, the spiking phase estimating operation $\mathcal{F}\{X_i\}$ in Equation (\ref{theta_cal}) is set as piecewise selection function as

\begin{equation}
	\varphi_i= \left\{\begin{array}{l}
	arctan(\frac{P_j}{Q_j}), \quad j = i,  \\
		0, \qquad\qquad\quad \ j \neq i.
	\end{array}\right.
\end{equation}

\subsubsection* {Robustness to reverse pixel noise and Gaussian noise}
Besides the effects of changing the whole background, we were interested in the capability of QS-SNN to handle other types of destabilization of images. For this purpose, we added reverse spike pixels and Gaussian noise to the MNIST and Fashion-MNIST images, and further tested the performance of QS-SNN in comparison with that of ANN and CNN. Reverse spike noise is created by randomly flipping image pixels to their reverse color and can be described as $Reverse(image[i])=1 -image[i]$. The position $i$ of the pixel to be flipped is randomly chosen, as shown in Figure \ref{image_reverse_pixels}(A) and (B).
The noisy images were encoded and processed in the same way as described in Algorithm S1. However, in the color-inverted experiment, all pixels of reverse degree $\theta_i$ were the same, resulting in the same change being applied to the whole image. By contrast, in the reverse pixel noise experiment, only a proportion of randomly chosen image pixels were changed; thus, every image pixel had a specific $\theta_i$ parameter. These image pixels are transferred to spike trains with heterogeneous phase $\varphi_i$. Specially in reverse pixel experiment we took the mean operation for $\mathcal{F}\{\cdot \}$ as the estimation phase:

\begin{equation}
	\varphi_i = \frac{1}{N}\sum_{j=1}^{N}arctan(\frac{P_j}{Q_j})
\end{equation}	 

Additive white Gaussian noise (AWGN) is commonly used to test system robustness. We also examined the performance of the proposed QS-SNN on AWGN MNIST and Fashion-MNIST images, as shown in Figure~\ref{gaussian_noise}(A) and (B).

In contrast to color-inverted noise, AWGN results in uncorrelated disturbances on the original image. We were interested in the robustness of our proposed method when faced with this challenging condition. The procedure used to process AWGN images was the same as that used in the reverse pixel noise experiment, except that the whole image phase was estimated using half the median operation:

\begin{equation}
	\varphi_{i} = M\{arctan(\frac{P_j}{Q_j})|j=1,...,N\}
\end{equation}

\section*{RESULTS}

\subsection*{Standard and color-inverted datasets experiment}
We constructed a three-layer QS-SNN with 500 hidden layer neurons and 10 output layer neurons. The structure of the experimental fully connected ANN was set to be the same for comparison. A simple CNN structure with three convolution layers and two pooling operations was used to determine the ability of different feature extraction methods to deal with inverted background images. We also tested VGG16, ResNet101, and DenseNet121 to investigate whether deeper structures could classify color-inverted images correctly. ANN, CNN, and QS-SNN were trained for 20 epochs with the Adam optimization method, and the learning rate was set to 0.001. VGG16, ResNet101, and DenseNet121 were trained for 400 epochs using stochastic gradient descent 
with learning rate 0.1, momentum 0.9, weight decay 5e-4, and learning rate decay 0.8 every 10 epochs. In the training phase, only the original image ($\theta=0$) was used; the testing phase used different color-inverted images ($\theta$ ranging from 0 to $\frac{\pi}{2}$). All results were obtained from the final epoch test step.

The results showed that the traditional fully connected ANN and convolution models struggled to handle huge changes in image properties such as background reversal, even when the spatial features of the image remained the same. Our proposed method showed much better performance than these traditional models (see Figures \ref{shift_exp}(C) and (D) and Tables S2, S3 for details). Significant performance degradation occurred when processing color-inverted images with ANN and CNN, and even deeper networks such as VGG16, ResNet101, and Densenet121 experienced problems with color-inverted image classification. By contrast, QS-SNN, although affected by a similar performance drop when images were made blurry ($\theta$ from 0 to $\frac{4\pi}{16}$), regained its ability when the images' backgrounds were inverted and the clarity was improved ($\theta$ from $\frac{4\pi}{16}$ to $\frac{8\pi}{16}$). When the image color was fully inverted ($\theta=\frac{8\pi}{16}$), QS-SNN retained the same accuracy as when classifying the original data ($\theta=0$) and correctly recognized color-inverted MNIST and Fashion-MNIST images.

\subsection* {Robustness to noise experiments}

\begin{figure}[!htb]
	\centering 
		\includegraphics[width=1.0\linewidth]{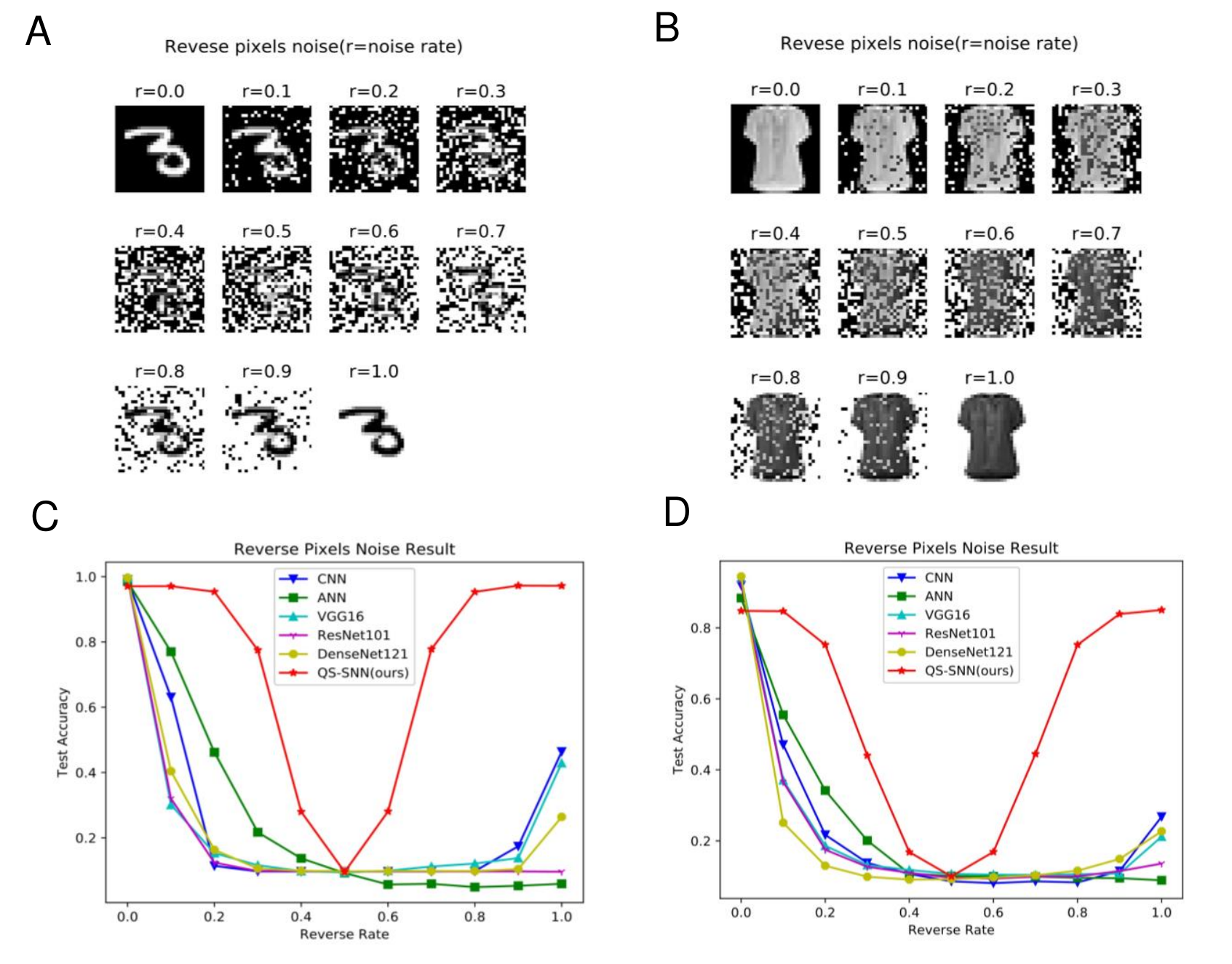}
	\caption{Reverse spike noise results. \textbf{(A)} Reverse spike noise MNIST. The possibility of pixel inversion is controlled by parameter $r$. When $r=0$, no noise is added, i.e., the image is original data. When $r=1.0$, all pixels are flipped. \textbf{(B)} Reverse spike noise Fashion-MNIST. \textbf{(C)} Classification of MNIST images with reverse noise. QS-SNN performed better compared with the inverted background experiment. \textbf{(D)} Classification of Fashion-MNIST images with reverse noise.}
	\label{image_reverse_pixels}
\end{figure}

\begin{figure}[!htb]
	\centering 
		\includegraphics[width=1.0\linewidth]{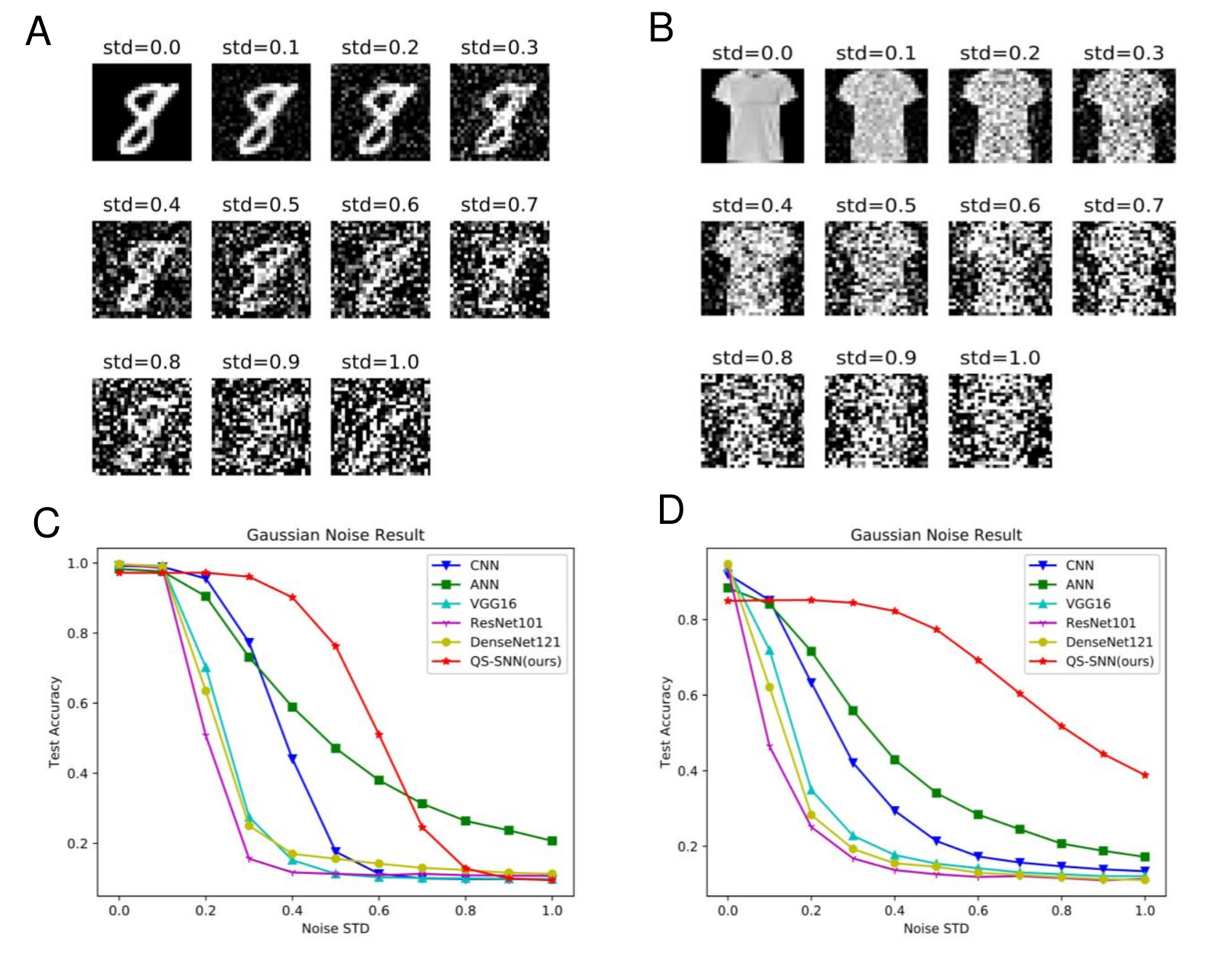}
	\caption{Gaussian noise image classification. \textbf{(A)} Additive white Gaussian noise on MNIST. The mean of Gaussian random noise was set to zero, and different $std$ values were used. \textbf{(B)} Additive white Gaussian noise on Fashion-MNIST. \textbf{(C)} Classification of MNIST images with Gaussian noise. Compared with other networks, QS-SNN showed much slower degeneration. \textbf{(D)} Classification of Fashion-MNIST images with Gaussian noise. QS-SNN performed even better compared with its results on MNIST.}
	\label{gaussian_noise}
\end{figure}

Compared with other state-of-the-art models, the performance of QS-SNN was closer to human vision capacity. As more flipped-pixel noise was added to the images ($r=0$ to $0.5$), they became increasingly difficult to recognize, as indicated by the left side of the `U'-curve for QS-SNN in Figure \ref{image_reverse_pixels}(C), (D). 
However, as more noise was added to the pixels, the image features became clear again. When $r=1.0$, with all pixels reversed, there was no conflict with the features of the original image when $r=0$. QS-SNN can exploit these conditions owing to its image superposition encoding method (Equation \ref{QS-SNN}), which is similar to the human vision system. As shown in Figure \ref{image_reverse_pixels}(C), (D) and Tables S4, S5, randomly inverting image pixels caused substantial performance degradation of ANN and CNN, as well as of the deep networks. On the contrary, the red `U'-shaped curve for QS-SNN indicated that it recovered its accuracy as the image's features became clear but the background was inverted ($r=0.6$ to $1.0$).

Gaussian noise influenced all networks significantly, with all methods showing a performance drop as noise (standard deviation; $std$) increased, as shown in Figure \ref{gaussian_noise}(C), (D) and Tables S6, S7. QS-SNN behaved more stably on the AWGN image processing task, with accuracies of 90.2\% and 82.3\% on the MNIST and Fashion-MNIST datasets, respectively, for $std=0.4$; by contrast, the other methods achieved no more than 60\% and 50\%, respectively. Images with $std=0.4$ are not very difficult for human vision to distinguish. Thus, by combining a brain-inspired spiking network with a quantum mechanism, we obtain a more robust approach to images with noise disturbance, similar to the performance of human vision.

\section*{Figure}

\section*{DISCUSSION}
This work aimed to integrate quantum theory with a biologically plausible SNN. Quantum image encoding and quantum superposition states were used for information representation, followed by processing with a spatial-temporal SNN. A time-convolution synapse was built to obtain neuron process phase information, and dendrite prediction with a proximal gradient method was used to train the QS-SNN. The proposed QS-SNN showed good performance on color-inverted image recognition tasks that were very challenging to other models. Compared with traditional ANN models, QS-SNN showed better generalization ability.

It is worth noting that the quantum brain hypothesis is quite controversial. Nevertheless, this paper does not aim to provide direct persuasive evidence for the quantum brain hypothesis but to explore novel information processing methods inspired by quantum information theory and brain spiking signal transmission.

\subsection*{Limitations of the study}
Our model was inspired by quantum image processing methods, in particular, quantum image superposition state presentation. The model and corresponding experiments were run on a classical computer and did not use any quantum hardware; thus, our work could not benefit from quantum computing. Artificial neurons can be reformed to run on quantum computers~\citep{schuld2014quest, cao2017quantum, mangini2020quantum}. Efforts to build a quantum spiking neuron are still at a preliminary stage~\citep{kristensen2019artificial}. Simulating spiking neuronal networks on a classical computer is hindered by heavy resource consumption and slow processing. 

Future work includes modifying both the quantum superposition encoding strategy and the SNN architecture to suit quantum computing better. In computational neuroscience research, neuronal spikes are typically generated with the Poisson process, which samples data from the binomial distribution. Quantum bits, also named qubits, are fundamental components in a quantum computer. A qubit can take the value of "0" or "1" with a certain probability, which is very similar to neuronal spikes. Thus, a set of qubits can encode all possible states of a spike train, as well as the quantum superposition images, in the quantum computer. Although it requires much effort to reconstruct spiking neural models suited for quantum computing, it is significant in neurology and artificial intelligence research to explore more quantum-inspired mechanisms to explain brain functions that traditional theories fail to.

\subsection*{Method details}
\subsubsection*{Generating background inverse image}
Different degree of background color inverse images $\mathinner{|I(\Theta)\rangle}$ using for experiment are generated according to the quantum superposition encoding. Here we describe the numerical form value of quantum image used to test model. Suppose the original image is represented by $X$ and the reversed image is $\bar{X}$. Parameter $\Theta$ controls the proportion of original image and reverse image in background color inverse images with
\begin{equation}
	I(\Theta)=Xcos(\Theta)+\bar{X}sin(\Theta)
	\label{color_invese_image_exp}
\end{equation}

\subsubsection*{Learning procedure of the QS-SNN algorithm}
Dendrite prediction~\citep{urbanczik2014learning} and proximal gradient methods are used for tuning of QS-SNN. In Equation (\ref{inputi}), $I_i$ is the teaching current, which is the integration of correct labels in $g_{E_i}(E_E-U_{1})$ and wrong labels in $g_{I_i}(E_I-U_{1})$. $E_E$ (8 mV) and $E_I$ (-8 mV) are the excitatory and inhibitory standard membrane potentials, respectively. The teaching current is injected to the soma of neurons in the output layer, generating added potential $V_{I_i}$  with membrane resistance $r_B$, as shown in Equation (\ref{outputmem}):
\begin{equation}
	\left\{\begin{array}{l}
		I^{ject}_{i}=g_{E_i}(E_E-U_{1})+g_{I_i}(E_I-U_{1}) \\
		g_{E_i}=
		\left\{
		\begin{array}{lr}
			1, \quad i=label, \\
			0, \quad i\neq label.
		\end{array}
		\right. \\
		g_{I_i}=
		\left\{
		\begin{array}{lr}
			0, \quad i=label, \\
			1, \quad i\neq label.
		\end{array}
		\right. \\
		V_{I_{i}} = r_B \cdot I^{ject}_{i},
	\end{array}\right.
	\label{inputi}
\end{equation}

\begin{equation}
	\tau_L\frac{dV_i^{o}(t)}{dt}=-V_i^o(t)+\frac{g_B}{g_L}(V_i^{o,b}-V_i^o(t))+V_{I_{i}}-V_i^o(t).
	\label{outputmem}
\end{equation}

Setting the left side of Equation (\ref{outputmem}) to zero and $V_i^{o}=V_{I_{i}}$, we get the steady state of somatic potentials $V_i^{o}$ and $V_i^{o,b}$ with $V_i^{o*}=g_B/(g_B+g_L)V_i^{o,b}$. The dendrite prediction rule defines the soma-dendrite error as Equation (\ref{lerror}):

\begin{equation}
	L=\frac{1}{2}\sum\limits_{i=0}^N\parallel r_{max}\sigma(V_i^{o})-r_{max}\sigma(V_i^{o*})\parallel^2.
	\label{lerror}
\end{equation}

Minimizing this error based on the differential chain rule, we obtain updated synaptic weights $w_{ij}^o$, as shown in Equations (\ref{diffw1}) and (\ref{diffb1}):

\begin{equation}
	\begin{aligned}
		\frac{\partial L}{\partial w_{ij}^o}&=\frac{\partial L}{\partial V_{i}^{o,b}}\frac{\partial V_{i}^{o,b}}{\partial w_{ij}^o}\\
		&=r_{max}\frac{g_B}{g_B+g_L}\left[\sigma(V_i^{o*})-\sigma(V_i^{o})\right]\sigma'(V_i^{o*})r_{j} \\
		&=\delta^{o}_{i}r_{j},
	\end{aligned}
	\label{diffw1}
\end{equation}

\begin{equation}
	\begin{aligned}
		\frac{\partial L}{\partial b_{o}^1}&=\frac{\partial L}{\partial V_{i}^{o,b}}\frac{\partial V_{i}^{o,b}}{\partial b_{i}^o}\\
		&=r_{max}\frac{g_B}{g_B+g_L}\left[\sigma(V_i^{o*})-\sigma(V_i^{o})\right]\sigma'(V_i^{o*})\\
		&=\delta^{o}_{i}.
	\end{aligned}
	\label{diffb1}
\end{equation}

Equation (\ref{update1}) shows the iterative updating of synaptic weights $w_{ij}^y$ and bias $b_{i}^y$:

\begin{equation}
	\left\{\begin{array}{l}
		w_{ij}^o \gets w_{ij}^o - \eta\frac{\partial L}{\partial w_{ij}^o} \\
		b_{i}^o \gets b_{i}^o - \eta\frac{\partial L}{\partial b_{i}^o}.
	\end{array}\right.
	\label{update1}
\end{equation}

For the hidden layer, error signal $\delta_{i}$ is passed from the previous layer, and neuron synapses $w_{ij}^h$ are adapted using Equations (\ref{diffw0}) and (\ref{diffb0}):

\begin{equation}
	\begin{aligned}
		\frac{\partial L}{\partial w_{ij}^h}&=\sum\limits_k\frac{\partial L}{\partial V_{k}^{o,b}} \frac{\partial V_{k}^{o,b}}{\partial r_{i}^h} \frac{\partial r_{i}^{h}}{\partial V_{i}^{h}}
		\frac{\partial V_{i}^{h}}{\partial V_{i}^{h, b}} \frac{\partial V_{i}^{h, b}}{\partial w_{ij}^h} \\
		&=\sum\limits_k\delta^{o}_{k} w_{ki}^{o} r_{max} \frac{g_B}{g_B+g_L} \sigma'(V_i^{h}) V_{j}^{PSP}  \\
		&=\sum\limits_k\delta^{o}_{k} \delta^{h}_{i} w_{ki}^{o} V_{j}^{PSP},
	\end{aligned}
	\label{diffw0}
\end{equation}

\begin{equation}
	\begin{aligned}
		\frac{\partial L}{\partial b_{i}^h}&=\sum\limits_k\frac{\partial L}{\partial V_{k}^{o,b}} \frac{\partial V_{k}^{o,b}}{\partial r_{i}^h} \frac{\partial r_{i}^{h}}{\partial V_{i}^{h}}
		\frac{\partial V_{i}^{h}}{\partial V_{i}^{h, b}} \frac{\partial V_{i}^{h, b}}{\partial b_{i}^h} \\
		&=\sum\limits_k\delta^{o}_{k} w_{ki}^{o} r_{max} \frac{g_B}{g_B+g_L} \sigma'(V_i^{h}) \\
		&=\sum\limits_k\delta^{o}_{k} \delta^{h}_{i} w_{ki}^{o}.
	\end{aligned}
	\label{diffb0}
\end{equation}

Equation (\ref{update0}) shows the iterative updating of synaptic weights $w_{ij}^h$ and bias $b_{i}^h$:

\begin{equation}
	\left\{\begin{array}{l}
		w_{ij}^h \gets w_{ij}^h - \eta\frac{\partial L}{\partial w_{ij}^h} \\
		b_{i}^h \gets b_{i}^h - \eta\frac{\partial L}{\partial b_{i}^h}.
	\end{array}\right.
	\label{update0}
\end{equation}

The training and test procedure for the QS-SNN model is shown in Algorithm~\ref{alg_QS-SNN}.

\begin{algorithm}[htbp]
	\footnotesize
	\caption{The learning procedure of QS-SNN.}
	\label{alg_QS-SNN}
	\newcommand{\INDState}{\State\hspace{\algorithmicindent}}
	\begin{algorithmic}
		\State {\bf 1.} Initialize weights $W_{j,i}$ with random uniform distribution, membrane potential states $V_i$, and other \\
		\quad related hyperparameters as in Table S1.	
		\State {\bf 2.} Start training procedure with only original images in training dataset,  $\theta_{i}=0$:	
		\INDState 2.1 Load training samples.
		\INDState 2.2 Construct quantum superposition state representations of images.
		\INDState 2.3 Input neuron spikes as Poisson process with spiking rate and phase time according to quantum \\ \quad \quad superposition image.
		\INDState 2.4 Process time-differential convolution to obtain dynamical updating of membrane potential of \\ \quad \quad post-synaptic neurons.
		\INDState 2.5 Update multi-layer membrane potential.
		\INDState 2.6 Train the QS-SNN with dendrite prediction and proximal gradient.
		\INDState 2.7 Select neurons in output layer with maximum spiking rate as the output class.
		\State {\bf 3.} Start test procedure using color-inverse images with different degree of color inversion from the test \\  \qquad dataset,  $\theta_{i}=0, \frac{\pi}{16}, \dots, \frac{8\pi}{16}$.
		\INDState 3.1 Load the test samples and transfer to spike trains as in steps 2.2 and 2.3.
		\INDState 3.2 Test the performance of the trained QS-SNN on color-inverse images.
		\INDState 3.3 Output the test performance.
	\end{algorithmic}
\end{algorithm}

\section*{Acknowledgments}
This study was supported by the new generation of artificial intelligence major project of the Ministry of Science and Technology of the People's Republic of China (Grant No. 2020AAA0104305),
the Strategic Priority Research Program of the Chinese Academy of Sciences (Grant No. XDB32070100) and the Beijing Municipal Commission of Science and Technology (Grant No. Z181100001518006).

\subsection*{Author contributions}
Y.S. wrote the code, performed the experiments, analyzed the data, and wrote the manuscript. Y.Z. proposed and supervised the project and contributed to writing the manuscript. T.Z. participated in helpful discussions and  contributed to writing the manuscript.

\subsection*{Declaration of interests}

The authors declare that they have no competing interests.

\setstretch{1.1}
\bibliography{QSNNRefs_LP}

\beginsupplement

\begin{figure}[thbp]
	\centering
	\includegraphics[width=14cm]{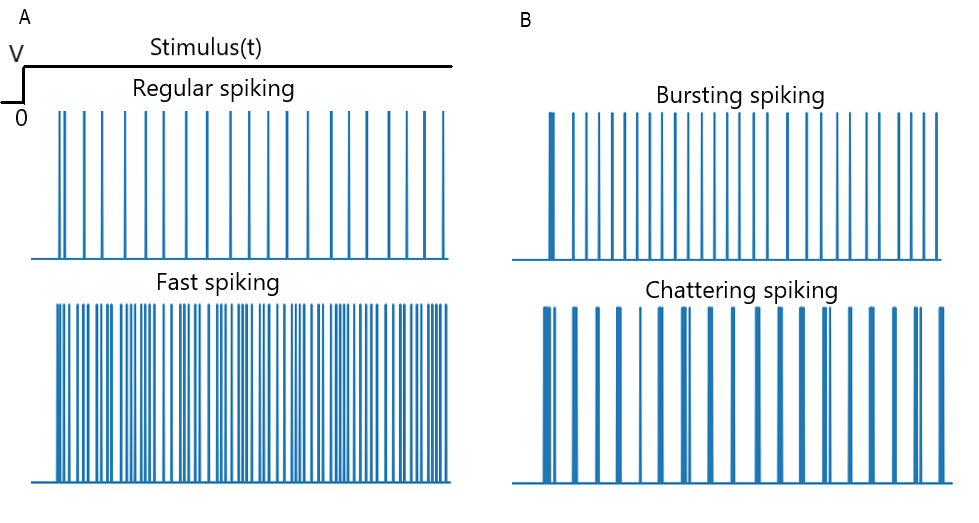}
	\caption{Spiking patterns with different fire rates or phases, Related to Figure 1. \textbf{(A)} Neurons fire at the same time but with different numbers of spikes generated per time unit. \textbf{(B)} Neurons fire the same number of spikes but the time of spiking is different.}
	\label{spikes}
\end{figure}

\begin{figure}[!htb]
	\centering	 
	\includegraphics[width=1.0\linewidth]{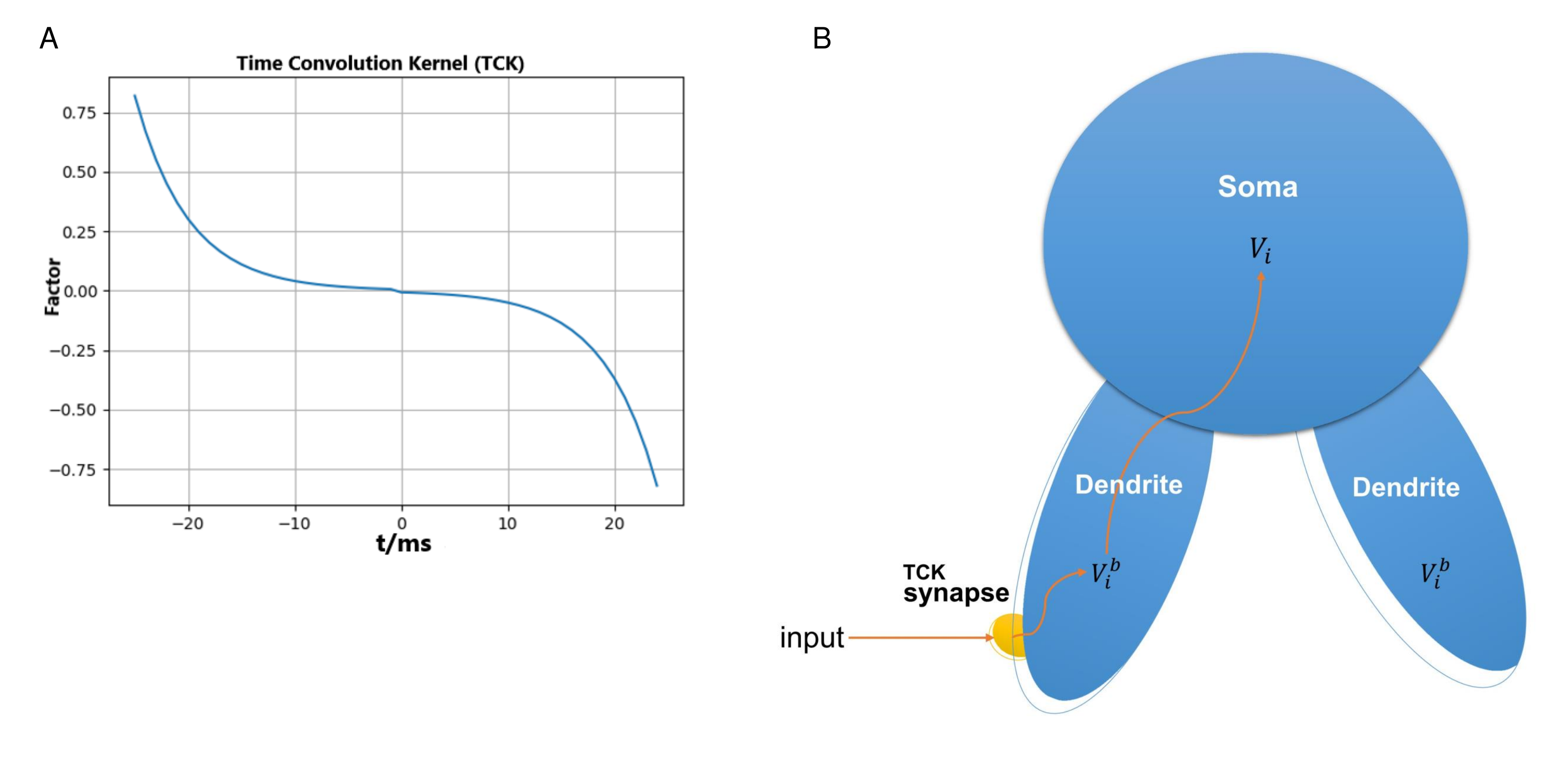}
	\caption{Two-compartment neuron mode,Related to STAR Methods. \textbf{(A)} Time convolution kernel. \textbf{(A)} Neuron with dendrite and soma compartments. Orange curve arrow shows information flowing from pre-synaptic input to dendrite potential $V_{i}^b$ and then integrated in soma potential $V_{i}$.}
	\label{TCK_mc_model}
\end{figure}

\begin{table}[htbp]
	\centering
	\caption{Hyperparameters of QS-SNN, Related to STAR Methods}
	\label{HP}
	\begin{tabular}{llll}
		\hline
		\hline
		Parameter & Value & Parameter & Value \\
		\hline
		$\tau$ & 4.0 & $r_{max}$ & 250 Hz\\
		$\tau_L$ & 10.0 ms & $T$ & 50 ms  \\
		$g_B$ & 0.6 nS  & $T_{sp}$ & 20 ms \\
		$g_L$ & 0.05 nS & $r_B$ & 1.0 n$\Omega$ \\
		\hline
		\hline
	\end{tabular}
\end{table}

\begin{table}[htbp]
	\centering
	\caption{MNIST color-inverted results  (accuracy percentages divided by 100), Related to Figure 4}
	\label{mnist_reverse_res_data}
	\begin{tabular}{|c|c|c|c|c|c|c|}
		\hline
		Algorithm &ANN &CNN &VGG16 &ResNet101 &DenseNet121 &QS-SNN \\
		\hline
		Structure &784-500-10 &784-c3p1c3p2c3-256-128-10 &- &- &-	&784-500-10	\\
		\hline
		$\theta=0$ & 0.978 & 0.992 & 0.996 & 0.996 & 0.997 & 0.971 \\
		\hline
		$\theta=\frac{\pi}{16}$ & 0.685 & 0.977 & 0.996 & 0.994 & 0.995 & 0.972 \\
		\hline
		$\theta=\frac{2\pi}{16}$ & 0.393 & 0.942 & 0.989 & 0.871 & 0.974 & 0.854 \\
		\hline
		$\theta=\frac{3\pi}{16}$ & 0.124 & 0.558 & 0.649 & 0.194 & 0.388 & 0.416\\
		\hline
		$\theta=\frac{4\pi}{16}$ & 0.097 & 0.114 & 0.114 & 0.098 & 0.114 & 0.089\\
		\hline
		$\theta=\frac{5\pi}{16}$ & 0.062 & 0.031 & 0.189 & 0.099 & 0.218 & \textbf{0.416}  \\
		\hline
		$\theta=\frac{6\pi}{16}$ & 0.054 & 0.138 & 0.458 & 0.103 & 0.372 & \textbf{0.923}\\
		\hline
		$\theta=\frac{7\pi}{16}$ & 0.059 & 0.381 & 0.504 & 0.102 & 0.362 & \textbf{0.970}\\
		\hline
		$\theta=\frac{8\pi}{16}$ & 0.062 & 0.469 & 0.488 & 0.099 & 0.350 & \textbf{0.973}\\
		\hline
	\end{tabular}
\end{table}

\begin{table}[ht]
	\centering
	\caption{Fashion-MNIST color-inverted results (accuracy percentages divided by 100), Related to Figure 4.}
	\label{fashion_reverse_res_data}
	\begin{tabular}{|c|c|c|c|c|c|c|}
		\hline
		Algorithm &ANN &CNN &VGG16 &ResNet101 &DenseNet121 &QS-SNN \\
		\hline
		Structure &784-500-10 &784-c3p1c3p2c3-256-128-10 &- &- &-	&784-500-10	\\
		\hline
		$\theta=0$ & 0.887 & 0.918 & 0.941 & 0.926 & 0.946 & 0.847 \\
		\hline
		$\theta=\frac{\pi}{16}$ & 0.417 & 0.812 & 0.602 & 0.304 & 0.718 & 0.826 \\
		\hline
		$\theta=\frac{2\pi}{16}$ & 0.194 & 0.693 & 0.437 & 0.126 & 0.524 & 0.539 \\
		\hline
		$\theta=\frac{3\pi}{16}$ & 0.100 & 0.494 & 0.308 & 0.101 & 0.311 & 0.224\\
		\hline
		$\theta=\frac{4\pi}{16}$ & 0.100 & 0.100 & 0.100 & 0.100 & 0.100 & 0.100\\
		\hline
		$\theta=\frac{5\pi}{16}$ & 0.100 & 0.131 & 0.169 & 0.100 & 0.130 & \textbf{0.224}  \\
		\hline
		$\theta=\frac{6\pi}{16}$ & 0.100 & 0.193 & 0.177 & 0.102 & 0.167 & \textbf{0.656}\\
		\hline
		$\theta=\frac{7\pi}{16}$ & 0.100 & 0.229 & 0.185 & 0.108 & 0.203 & \textbf{0.827}\\
		\hline
		$\theta=\frac{8\pi}{16}$ & 0.098 & 0.244 & 0.190 & 0.120 & 0.249 & \textbf{0.847}\\
		\hline
	\end{tabular}
\end{table}

\begin{table}[ht]
	\centering
	\caption{MNIST reverse-pixels noise results (accuracy percentages divided by 100), Related to Figure 5.}
	\label{mnist_reverse_pix_res_data}
	\begin{tabular}{|c|c|c|c|c|c|c|}
		\hline
		Algorithm &ANN &CNN &VGG16 &ResNet101 &DenseNet121 &QS-SNN \\
		\hline
		Structure &784-500-10 &784-c3p1c3p2c3-256-128-10 &- &- &-	&784-500-10	\\
		\hline
		$r=0$ & 0.985 & 0.991 & 0.996 & 0.996 & 0.997 & 0.971 \\
		\hline
		$r=0.1$ & 0.77 & 0.631 & 0.301 & 0.320 & 0.404 & \textbf{0.971} \\
		\hline
		$r=0.2$ & 0.462 & 0.114 & 0.153 & 0.124 & 0.163 & \textbf{0.954} \\
		\hline
		$r=0.3$ & 0.217 & 0.097 & 0.116 & 0.098 & 0.106 & \textbf{0.775}\\
		\hline
		$r=0.4$ & 0.137 & 0.097 & 0.098 & 0.098 & 0.099 & 0.280\\
		\hline
		$r=0.5$ & 0.092 & 0.097 & 0.094 & 0.097 & 0.097 & 0.095  \\
		\hline
		$r=0.6$ & 0.057 & 0.097 & 0.099 & 0.097 & 0.098 & 0.281\\
		\hline
		$r=0.7$ & 0.059 & 0.097 & 0.112 & 0.097 & 0.098 & \textbf{0.778} \\
		\hline
		$r=0.8$ & 0.049 & 0.098 & 0.121 & 0.097 & 0.098 & \textbf{0.953}\\
		\hline
		$r=0.9$ & 0.053 & 0.174 & 0.138 & 0.097 & 0.104 & \textbf{0.972}\\
		\hline
		$r=1.0$ & 0.059 & 0.464 & 0.429 & 0.096 & 0.264 & \textbf{0.972}\\
		\hline
	\end{tabular}
\end{table}

\begin{table}[ht]
	\centering
	\caption{Fashion-MNIST reverse-pixel noise results (accuracy percentages divided by 100), Related to Figure 5.}
	\label{fashion_mnist_reverse_pix_res_data}
	\begin{tabular}{|c|c|c|c|c|c|c|}
		\hline
		Algorithm &ANN &CNN &VGG16 &ResNet101 &DenseNet121 &QS-SNN \\
		\hline
		Structure &784-500-10 &784-c3p1c3p2c3-256-128-10 &- &- &-	&784-500-10	\\
		\hline
		$r=0$ & 0.884 & 0.920 & 0.94 & 0.927 & 0.945 & 0.848 \\
		\hline
		$r=0.1$ & 0.555 & 0.471 & 0.371 & 0.366 & 0.251 & \textbf{0.847} \\
		\hline
		$r=0.2$ & 0.342 & 0.217 & 0.186 & 0.175 & 0.130 & \textbf{0.752} \\
		\hline
		$r=0.3$ & 0.201 & 0.138 & 0.132 & 0.127 & 0.099 & \textbf{0.441}\\
		\hline
		$r=0.4$ & 0.108 & 0.107 & 0.118 & 0.110 & 0.091 & 0.169\\
		\hline
		$r=0.5$ & 0.101 & 0.086 & 0.107 & 0.096 & 0.092 & 0.100 \\
		\hline
		$r=0.6$ & 0.100 & 0.081 & 0.105 & 0.094 & 0.097 & 0.169\\
		\hline
		$r=0.7$ & 0.099 & 0.086 & 0.104 & 0.099 & 0.103 & \textbf{0.445}\\
		\hline
		$r=0.8$ & 0.096 & 0.083 & 0.106 & 0.100 & 0.116 & \textbf{0.752}\\
		\hline
		$r=0.9$ & 0.095 & 0.115 & 0.108 & 0.115 & 0.149 & \textbf{0.839}\\
		\hline
		$r=1.0$ & 0.089 & 0.268 & 0.212 & 0.136 & 0.227 & \textbf{0.850}\\
		\hline
	\end{tabular}
\end{table}

\begin{table}[ht]
	\centering
	\caption{MNIST Gaussian noise results (accuracy percentages divided by 100), Related to Figure 6.}
	\label{mnist_gaussian_data}
	\begin{tabular}{|c|c|c|c|c|c|c|}
		\hline
		Algorithm &ANN &CNN &VGG16 &ResNet101 &DenseNet121 &QS-SNN \\
		\hline
		Structure &784-500-10 &784-c3p1c3p2c3-256-128-10 &- &- &-	&784-500-10	\\
		\hline
		$std=0$ & 0.983 & 0.992 & 0.996 & 0.995 & 0.997 & 0.972 \\
		\hline
		$std=0.1$ & 0.976 & 0.989 & 0.992 & 0.986 & 0.990 & 0.972 \\
		\hline
		$std=0.2$ & 0.905 & 0.956 & 0.702 & 0.507 & 0.635 & \textbf{0.972} \\
		\hline
		$std=0.3$ & 0.731 & 0.773 & 0.275 & 0.156 & 0.250 & \textbf{0.961} \\
		\hline
		$std=0.4$ & 0.589 & 0.441 & 0.152 & 0.117 & 0.156 & \textbf{0.902} \\
		\hline
		$std=0.5$ & 0.471 & 0.176 & 0.113 & 0.113 & 0.142 & \textbf{0.763} \\
		\hline
		$std=0.6$ & 0.380 & 0.113 & 0.103 & 0.109 & 0.130 & \textbf{0.510} \\
		\hline
		$std=0.7$ & 0.313 & 0.100 & 0.101 & 0.113 & 0.124 & 0.245 \\
		\hline
		$std=0.8$ & 0.264 & 0.098 & 0.099 & 0.108 & 0.116 & 0.128 \\
		\hline
		$std=0.9$ & 0.237 & 0.098 & 0.099 & 0.108 & 0.116 & 0.100 \\
		\hline
		$std=1.0$ & 0.207 & 0.097 & 0.098 & 0.107 & 0.113 & 0.095 \\
		\hline
	\end{tabular}
\end{table}

\begin{table}[ht]
	\centering
	\caption{Fashion-MNIST Gaussian noise results (accuracy percentages divided by 100), Related to Figure 6.}
	\label{fashion_mnist_gaussian_data}
	\begin{tabular}{|c|c|c|c|c|c|c|}
		\hline
		Algorithm &ANN &CNN &VGG16 &ResNet101 &DenseNet121 &QS-SNN \\
		\hline
		Structure &784-500-10 &784-c3p1c3p2c3-256-128-10 &- &- &-	&784-500-10	\\
		\hline
		$std=0$ & 0.884 & 0.920 & 0.938 & 0.934 & 0.947 & 0.850 \\
		\hline
		$std=0.1$ & 0.841 & 0.851 & 0.719 & 0.464 & 0.621 & 0.851 \\
		\hline
		$std=0.2$ & 0.716 & 0.633 & 0.349 & 0.251 & 0.283 & \textbf{0.852} \\
		\hline
		$std=0.3$ & 0.559 & 0.421 & 0.228 & 0.168 & 0.193 & \textbf{0.845} \\
		\hline
		$std=0.4$ & 0.429 & 0.294 & 0.177 & 0.137 & 0.155 & \textbf{0.823}\\
		\hline
		$std=0.5$ & 0.341 & 0.214 & 0.154 & 0.126 & 0.146 & \textbf{0.774} \\
		\hline
		$std=0.6$ & 0.284 & 0.173 & 0.142 & 0.119 & 0.13 & \textbf{0.692} \\
		\hline
		$std=0.7$ & 0.245 & 0.157 & 0.131 & 0.121 & 0.124 & \textbf{0.604} \\
		\hline
		$std=0.8$ & 0.207 & 0.147 & 0.126 & 0.116 & 0.118 & \textbf{0.518} \\
		\hline
		$std=0.9$ & 0.188 & 0.139 & 0.121 & 0.110 & 0.113 & \textbf{0.444} \\
		\hline
		$std=1.0$ & 0.172 & 0.134 & 0.121 & 0.114 & 0.111 & \textbf{0.388} \\
		\hline
	\end{tabular}
\end{table}

\end{document}